\pdfoutput=1
\documentclass{article}

\usepackage{afterpage}
\usepackage{arxiv}

\usepackage[utf8]{inputenc} 
\usepackage[T1]{fontenc}    
\usepackage{hyperref}       
\usepackage{url}            
\usepackage{booktabs}       
\usepackage{amsfonts}       
\usepackage{nicefrac}       
\usepackage{microtype}      
\usepackage{lipsum}
\usepackage{graphicx}
\usepackage{times}
\usepackage{soul}
\usepackage{url}
\usepackage{graphicx}
\usepackage{amsthm}

\usepackage{amsmath}
\usepackage{amsthm}
\usepackage{booktabs}
\usepackage{algorithm}
\usepackage{algpseudocode}
\usepackage{algorithmicx}
\usepackage[switch]{lineno}
\usepackage{natbib}

\usepackage{mathtools}
\usepackage{amssymb}
\usepackage{xcolor}
\usepackage{multirow}
\usepackage{pifont}
\usepackage{threeparttable}
\graphicspath{ {./images/} }
\usepackage{subcaption}

\title{GEM: A Scale-Aware and Distribution-Sensitive Sparse Fine-Tuning Framework
for Effective Downstream Adaptation}

\author{
Sungmin Kang \\
University of Southern California\\
\texttt{kangsung@usc.edu} 
\And
Jisoo Kim \\
Inha University\\
\texttt{starprin3@inha.edu} 
\AND
Salman Avestimehr\\
University of Southern California\\
\texttt{avestime@usc.edu}
\And
Sunwoo Lee \\
Inha University\\
\texttt{sunwool@inha.ac.kr} \\
}

\begin{document}
\maketitle
\begin{abstract}
Parameter-efficient fine-tuning (PEFT) has become a popular way to adapt large pre-trained models to new tasks. Most PEFT methods update only a small subset of parameters while freezing the rest, avoiding redundant computation. As they maximize the absolute size of the updates without regard to the parameters’ original scale, the resulting changes in model behavior can be minimal. In contrast, we maximize updates relative to each parameter’s scale, yielding more meaningful downstream adaptation. We propose \textbf{G}radient-to-Weight Ratio and \textbf{E}ntropy-guided \textbf{M}asking (GEM), a parameter scale-aware, distribution-sensitive sparse fine-tuning framework. GEM prioritizes parameters whose updates are significant in proportion to their initial pre-trained values. It also adaptively determines how many parameters to tune at each layer based on the entropy of parameter values, thereby making the most effective use of the computational budget in PEFT. Our empirical study demonstrates the efficacy of GEM on both general-domain tasks (GLUE and SuperGLUE) and domain-specific tasks (GSM8k and MBPP), achieving up to a $1.6\%$ improvement in fine-tuning accuracy over full fine-tuning while updating only $0.1\%$ of model parameters.
\end{abstract}

\section{Introduction} \label{sec:intro}
With the success of large-scale pre-trained models~\cite{brown2020languagemodelsfewshotlearners, zhang2022optopenpretrainedtransformer, touvron2023llamaopenefficientfoundation, gunasekar2023textbooksneed}, fine-tuning has been an efficient yet effective way of adapting large models to downstream tasks using fewer training data. Despite the success, fine-tuning the entire large-scale model demands intensive resource and time, which poses significant challenges to adapt the model to multiple target tasks. To reduce the computational burden, parameter-efficient fine-tuning (PEFT) methods have been proposed with the aim to adapt models by updating only a small subset of parameters while keeping the majority of weights fixed.

Several lines of research have been proposed to address the high cost of fine-tuning, such as Low-rank Adaptation (LoRA) \cite{hu2021loralowrankadaptationlarge}, Adapter \cite{houlsby2019parameterefficienttransferlearningnlp}, and prompt tuning \cite{li2021prefixtuningoptimizingcontinuousprompts}. Parameter masking-based sparse fine-tuning methods \cite{zhao2020maskingefficientalternativefinetuning} have also been proposed, showing that most parameters can be frozen, with only a small portion being fine-tuned to achieve comparable accuracy to full fine-tuning. These methods construct masks by estimating the importance of individual parameters and selecting those with the highest scores, using criteria such as gradient magnitude \cite{li2025enhancinglargelanguagemodel, zhang2024gradientbasedparameterselectionefficient}, Fisher matrix \cite{agarwal2025stepbystepunmaskingparameterefficientfinetuning, sung2021trainingneuralnetworksfixed, das2023unifiedlowresourcesequencelabeling}, or Hessian-based sensitivity measures \cite{xu2025adaptiveparameterefficientfinetuninghessianinformed}. 
However, existing approaches overlook the scale of pre-trained parameter weights during parameter selection.

A common goal of PEFT is to maximize \textit{downstream adaptation} within a constrained budget.
Here, we define \textit{downstream adaptation} not as the absolute magnitude of parameter updates, but as the extent to which parameters change \textit{relative to their pre-trained values}.
In other words, a parameter that undergoes a large proportional shift from its initial value is considered to contribute more significantly to task-specific adaptation, even if its absolute update is small.
Therefore, maximizing downstream adaptation under a limited budget entails identifying and updating parameters that exhibit the greatest relative change.
This motivates our focus on relative parameter change $\lvert \Delta w \rvert / \lvert w \rvert$ as a principled criterion for identifying the most adaptable parameters under tight tuning budgets.
In this work, we propose GEM, a general PEFT framework built upon a novel parameter prioritization metric and an entropy-guided parameter selection method.
The main principle of our framework is to select a small portion of parameters that have the most critical impact on model behavior and fine-tune only those parameters efficiently.
First, we investigate the impact of a parameter selection strategy based on the gradient magnitude relative to the parameter values.
We empirically explore and demonstrate why the gradient-to-weight ratio identifies important parameters more effectively than the gradient magnitude alone.
Second, we investigate how to exploit entropy information at each network layer to determine the number of parameters to be tuned per layer.
Our study empirically proves that entropy is a useful metric for identifying layers with an irregular distribution of parameter importance.
We can safely freeze more parameters in layers with irregular distributions, expecting a greater reduction in computational cost while maintaining fine-tuning accuracy.

\begin{table*}[t]
\centering
\footnotesize
\begin{threeparttable}
\resizebox{\textwidth}{!}{%
\begin{tabular}{lcccccccc}
\toprule
\textbf{Method} &
\shortstack[c]{\textbf{Model}\\\textbf{Agnostic}} &
\shortstack[c]{\textbf{No Extra}\\\textbf{Train Param}} &
\shortstack[c]{\textbf{No Extra}\\\textbf{Infer Param}} &
\shortstack[c]{\textbf{Param-wise}\\\textbf{Importance}} &
\shortstack[c]{\textbf{Layer-wise}\\\textbf{Importance}} &
\shortstack[c]{\textbf{Parameter}\\\textbf{Scale-aware}} &
\shortstack[c]{\textbf{Tuning Budget}\\\textbf{Controllable}} &
\shortstack[c]{\textbf{Static}\\\textbf{Mask}} \\
\midrule
LoRA              & \ding{55} & \ding{55} & \ding{51} & \ding{55} & \ding{55} & \ding{55} & \ding{55} & \ding{51} \\
BitFit            & \ding{51} & \ding{51} & \ding{51} & \ding{55} & \ding{55} & \ding{55} & \ding{55} & \ding{51} \\
Adapter           & \ding{55} & \ding{55} & \ding{55} & \ding{55} & \ding{55} & \ding{55} & \ding{55} & \ding{51} \\
AdaLoRA           & \ding{55} & \ding{55} & \ding{51} & \ding{51} & \ding{51} & \ding{55} & \ding{55} & \ding{55} \\
Random Mask       & \ding{51} & \ding{51} & \ding{51} & \ding{55} & \ding{55} & \ding{55} & \ding{51} & \ding{51} \\
Top-Grad Mask     & \ding{51} & \ding{51} & \ding{51} & \ding{51} & \ding{51}\tnote{*} & \ding{55} & \ding{51} & \ding{51} \\
\textbf{GEM (ours)} & \ding{51} & \ding{51} & \ding{51} & \ding{51} & \ding{51} & \ding{51} & \ding{51} & \ding{51} \\
\bottomrule
\end{tabular}%
}
\begin{tablenotes}
\footnotesize
\item[*] Some implementations of Top-Grad Masking optionally consider per-layer allocation.
\end{tablenotes}
\end{threeparttable}
\caption{Comparison of GEM with representative PEFT baselines. GEM is a versatile method that combines scale-awareness, hierarchical selectivity, and model-agnostic design, while being static and parameter-efficient.}
\end{table*}

To evaluate the performance of GEM, we conduct benchmark experiments on several popular datasets, including RTE, SST-2, WiC, BoolQ, MultiRC, CoPA, and SQuAD.
We directly compare our proposed method to a variety of SOTA PEFT methods such as LoRA, BitFit, Adapter, AdaLoRA, random masking, and top gradient masking.
Our extensive experiments consistently demonstrate that the gradient-to-weight ratio and the entropy-aware parameter selection strategy achieve the best fine-tuning accuracy while significantly reducing the computational cost.
We also present a few ablation study results to further evaluate the efficacy of our method.
Based on our observations and analyses, we conclude that GEM is a highly efficient PEFT framework that can be readily applied to a wide range of real-world LLM-based applications.

Our main contributions are summarized as follows:
\begin{itemize}
    \item We propose a novel parameter prioritization method that maximizes weight changes relative to their original scale and analyze its impact on fine-tuning performance.
    \item We propose an entropy-guided, layer-wise parameter-allocation method that dynamically selects the number of tunable parameters in each layer, and we analyze how exploiting both the magnitude and distribution of learning signals maximizes fine-tuning performance.
    \item Building on these two methods, we introduce a general PEFT framework, GEM, and empirically validate its effectiveness through extensive experiments across diverse tasks, including both general-domain and domain-specific benchmarks.
\end{itemize}
\section{Related Works} \label{sec:related}
We first group existing PEFT methods into three categories: (1) reparameterization-based methods, which introduce tunable modules or embeddings into a frozen model; (2) sparsification and masking-based methods, which selectively update a subset of important parameters; and (3) quantization- and pruning-based methods, which compress the model to reduce memory or computational cost during fine-tuning. Some recent approaches combine ideas from multiple categories to improve performance or efficiency.

\subsection{Reparameterization-based Methods.}
A key family of PEFT techniques reparameterizes pre-trained models by modifying internal parameterizations, either by injecting lightweight modules, or prepending tunable inputs, while keeping the original model weights frozen.
Low-Rank Adaptation (LoRA) \cite{hu2021loralowrankadaptationlarge} is a popular PEFT method, which utilizes a low-rank decomposition to efficiently fine-tune models with minimal parameter overhead.
There are adaptive rank allocation methods such as AdaLoRA~\cite{zhang2023adaloraadaptivebudgetallocation} and ALoRA~\cite{liu2024aloraallocatinglowrankadaptation}.
GoRA~\cite{he2025goragradientdrivenadaptivelow} and SoRA~\cite{ding2023sparselowrankadaptationpretrained} utilize gradient-based signals to control rank allocation.
LoRA-FA~\cite{zhang2023lorafamemoryefficientlowrankadaptation}, VeRA~\cite{kopiczko2024veravectorbasedrandommatrix}, and S-LoRA~\cite{ev2024xora} enhance memory efficiency by limiting the number, or the extent, of trainable parameters.
Introduced by Houlsby et al.~\cite{houlsby2019parameterefficienttransferlearningnlp}, adapters are lightweight modules inserted within transformer layers to enable task-specific adaptation, significantly reducing the number of tunable parameters.
Parallel adapters \cite{pfeiffer2020adapterhubframeworkadaptingtransformers} allow adapters to be inserted alongside the main path for better gradient flow, while AdapterFusion \cite{pfeiffer2021adapterfusionnondestructivetaskcomposition} enables multi-task adaptation by learning to combine knowledge from multiple adapters. Compacter~\cite{mahabadi2021compacterefficientlowrankhypercomplex} factorizes adapter weights into shared low-rank components.

\subsection{Sparsification and Masking-based Methods.}
Motivated by the lottery ticket hypothesis~\cite{frankle2019lotterytickethypothesisfinding} and structured pruning techniques~\cite{guo2021parameterefficienttransferlearningdiff}, sparsification and masking have been widely applied to PEFT. 
These methods identify a subset of parameters based on importance measures and fine-tune the subset only.
Another line of work leverages gradient magnitude as a proxy for parameter importance, either by pre-selecting weights with large gradients or dynamically masking them during training~\cite{zhang2024gradientbasedparameterselectionefficient, li2025enhancinglargelanguagemodel, song2024sparsefinetuningpretrainedlarge}.
Child-Tuning~\cite{xu2021raisechildlargelanguage} restricts updates to a subnetwork selected either randomly or based on accumulated gradient norms.
More recently, random masking ~\cite{xu2024randommaskingfindswinning}, which selects a random subset to update, has also shown surprisingly competitive performance.

\subsection{Quantization- and Pruning-based Methods.}
Quantization-based methods lower the precision of model weights to reduce memory footprint and accelerate computation. For example, QLoRA~\cite{dettmers2023qloraefficientfinetuningquantized} enables 4-bit fine-tuning of large language models with minimal performance degradation, while GPTQ~\cite{frantar2023gptqaccurateposttrainingquantization} and AWQ~\cite{lin2024awqactivationawareweightquantization} provide post-training quantization techniques for compressing LLMs efficiently. These methods allow parameter-efficient tuning of LLMs under severe memory constraints.
Pruning-based methods explicitly remove a subset of parameters to construct a sparse subnetwork, reducing training cost and inference overhead. DiffPruning~\cite{guo2021parameterefficienttransferlearningdiff} learns a binary mask that selects only a small fraction of weights to update during fine-tuning, while keeping the rest frozen. Light-PEFT~\cite{gu2024lightpeftlighteningparameterefficientfinetuning} adopts an early-stage pruning strategy that estimates parameter importance in the initial training phase, enabling the removal of less critical components before full adaptation.

\section{Method} \label{sec:method}

\begin{figure*}[t]
    \centering
    \includegraphics[width=1\textwidth]{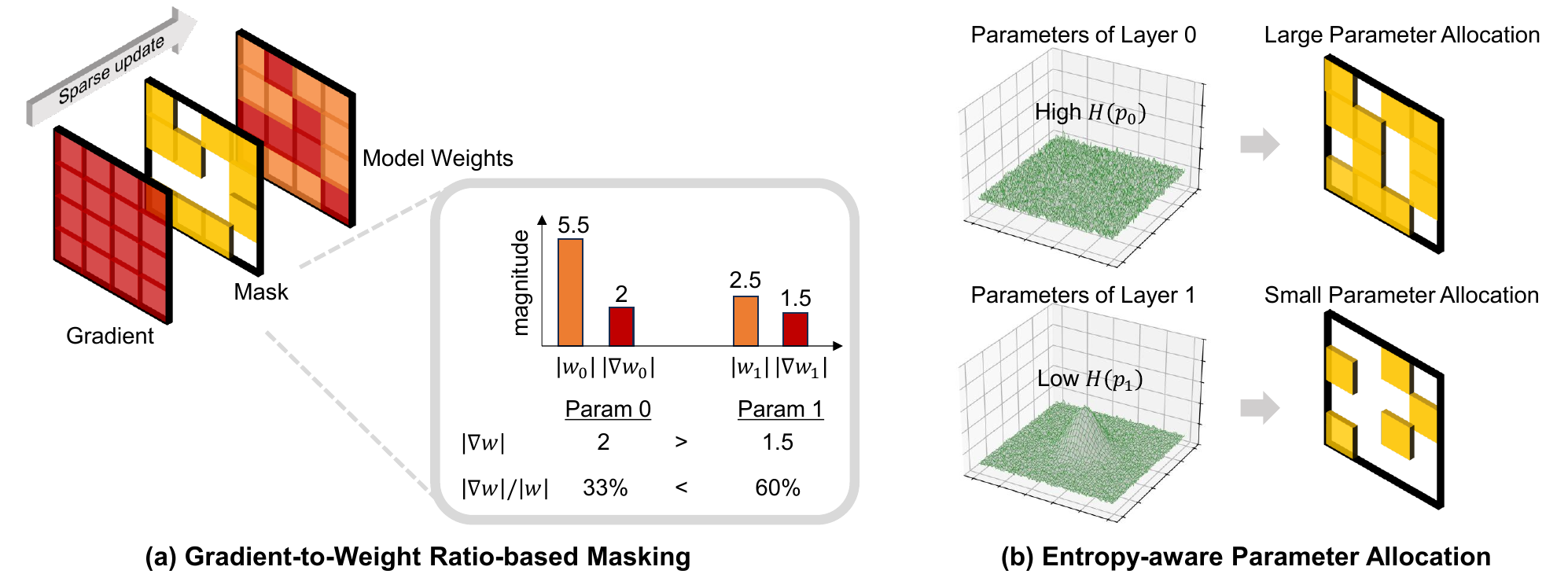}
    \caption{
    Schematic illustration of GEM framework's key components.
    (a) We prioritize parameters that maximize the relative weight change, rather than the absolute weight change commonly used in prior methods.
    This figure illustrates a simple example in which the important parameter varies with the metric used. 
    While existing methods select parameter 0 based on its larger gradient magnitude, our method selects parameter 1 due to its larger gradient relative to its original weight.
    (b) The number of tunable parameters in each layer is determined by the degree to which the learning signal is concentrated.
    We quantify layer-wise allocations using entropy-based importance scores.
    }
    \label{fig:GEM_framework}
\end{figure*}

In this section, we propose a binary masking framework for selecting tunable parameters to achieve accurate and parameter-efficient fine-tuning.
Our framework comprises two key components: (i) parameter prioritization based on the gradient-to-weight ratio, and (ii) entropy-guided layer-wise parameter selection. 
Based on these two methods, we construct a general PEFT framework, GEM. 
Figure~\ref{fig:GEM_framework} shows an overview of GEM, illustrating how it selects the most critical and minimal subset of parameters to tune.

\subsection{Gradient-to-Weight Ratio for Parameter Prioritization}
\label{subsection:Gradient-to-Weight-Ratio-for-Parameter-Selection}

\subsubsection{Parameter Prioritization using a Scale-Invariant Signal.}
Consider a standard first-order update rule: $w_{t+1} = w_t - \eta \cdot \nabla_{w} \mathcal{L}$, where $\eta$ is the learning rate. The weight update magnitude is $|\Delta w| = \eta \cdot |\nabla_{w} \mathcal{L}|$, and its relative scale with respect to the parameter is given by
\begin{equation}
\frac{|\Delta w|}{|w|} = \eta \cdot \frac{|\nabla_{w} \mathcal{L}|}{|w|}.
\label{eq:relative-update}
\end{equation}

Motivated by the analysis above, we define a scale-invariant signal for parameter prioritization, referred to as the gradient-to-weight ratio (GWR), as
\begin{equation}
\rho^{(i)} := \left| \frac{\nabla_{w^{(i)}} \mathcal{L}}{w^{(i)}} \right|,
\label{eq:gradient-to-weight}
\end{equation}
where \( w^{(i)} \) denotes the scalar \( i \)-th element of the weight tensor $W$.
We consider parameters with a high $\rho$ value to be critical in the fine-tuning process.
Even if two parameters have gradients of comparable magnitude, the parameter with smaller weight will experience a larger relative update, potentially causing greater change in the model’s behavior.
This suggests that relying solely on gradient magnitude may overlook the relative scale of updates, while the gradient-to-weight ratio provides a more accurate and scale-aware measure of parameter influence.
Note that this signal is naturally available during training, as both the numerator and the denominator are computed at every iteration, incurring almost no additional computational cost.

\subsubsection{Theoretical Interpretation of Gradient-to-Weight Ratio.}
The gradient-to-weight ratio can be interpreted as a good metric for quantifying parameter importance.
Let us begin with a first-order Taylor approximation of the loss $\mathcal{L}(w)$:
\begin{align}
\mathcal{L}(w + \Delta w) 
&\approx \mathcal{L}(w) + \nabla_w \mathcal{L}(w)^T \Delta w \\
\quad 
\Delta \mathcal{L} 
&\approx \nabla_w \mathcal{L}(w)^T \Delta w
\end{align}

Assuming a gradient descent update \( \Delta w = -\eta \nabla_w \mathcal{L}(w) \), the change in loss becomes:
\[
\Delta \mathcal{L} \approx -\eta \|\nabla_w \mathcal{L}(w)\|^2 
\hspace{0.1cm} \Rightarrow \hspace{0.1cm} 
\frac{|\Delta \mathcal{L}|}{\|\nabla_w \mathcal{L}(w)\|} \approx \eta \|\nabla_w \mathcal{L}(w)\|.
\]

This shows that the quantity \( |\Delta \mathcal{L}| / \|\nabla_w \mathcal{L}(w)\| \) characterizes the reduction in loss per unit of gradient norm when taking a step along the negative gradient direction. In other words, it measures how efficiently the loss is decreased relative to the local steepness of the loss landscape at the current parameter setting.

Meanwhile, the norm of the update satisfies \( \|\Delta w\| = \eta \|\nabla_w \mathcal{L}(w)\| \), leading to:
\[
\frac{|\Delta \mathcal{L}| / \|\nabla_w \mathcal{L}(w)\|}{\|w\|} 
\approx 
\frac{\|\Delta w\|}{\|w\|}.
\]

Thus, the left-hand side can be interpreted as a scale-invariant measure of effective loss reduction, quantifying optimization progress relative to both gradient magnitude and parameter scale.
The right-hand side represents the relative parameter update, and this equivalence highlights a direct connection between the extent of loss reduction and the magnitude of parameter changes relative to their scale.
Based on this, we conclude that the gradient-to-weight ratio effectively captures how significantly each parameter contributes to reducing the loss.

\begin{figure}[t]
    \centering
    \begin{subfigure}[t]{0.45\linewidth}
        \centering
        \includegraphics[width=0.9\linewidth]{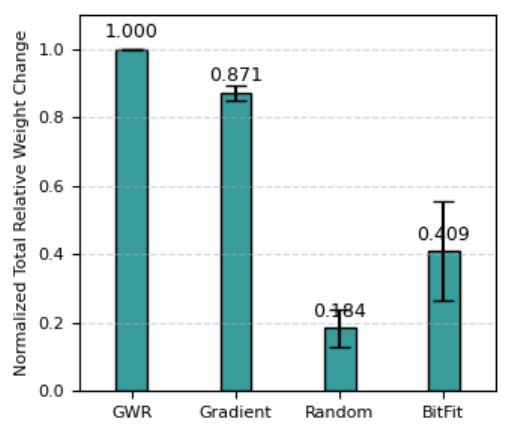}
        \caption{Relative weight change}
        \label{fig:wic_rel_change}
    \end{subfigure}%
    \hspace{0.04\linewidth}
    \begin{subfigure}[t]{0.45\linewidth}
        \centering
        \includegraphics[width=0.9\linewidth]{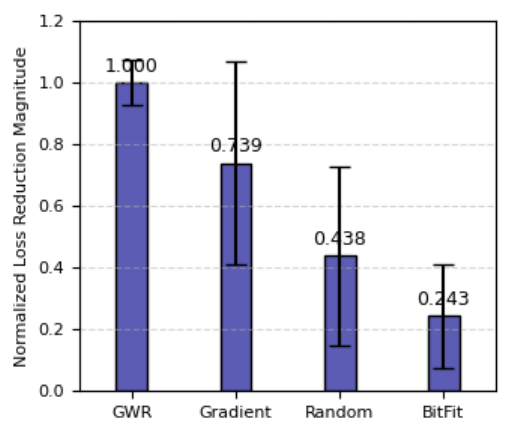}
        \caption{Loss reduction}
        \label{fig:wic_loss_reduction}
    \end{subfigure}
    \caption{
    \textbf{Gradient-to-weight ratio enables identifying critical parameters by prioritizing updates that better align with loss reduction.} 
    We fine-tune OPT-125M on WIC for 10 epochs, testing four masking strategies over three seeds. Results are normalized. 
    Plot (a) shows total deviation from the pre-trained weights; plot (b) shows corresponding loss reduction. 
    Gradient-to-weight masking produces both the largest relative weight shifts and the greatest loss drop.
    }
    \label{fig:relative_weight_change}
\end{figure}

\subsubsection{Empirical Interpretation of Gradient-to-Weight Ratio.}
Beyond our theoretical understanding, we empirically investigate the effectiveness of the gradient-to-weight ratio as a metric for quantifying parameter importance. To assess the impact of parameter selection methods on fine-tuning efficiency, we evaluate two metrics:  
(1) total relative weight change with respect to the initial pre-trained weights, measured as \( \left\| \frac{w_t - w_0}{w_0} \right\| \), and  
(2) the magnitude of loss reduction, computed as \( |\Delta \mathcal{L}| \approx \left| -\nabla_{w} \mathcal{L}_0 \cdot \left( w_t - w_0 \right) \right| \),  
where \( w_0 \) and \( w_t \) denote the initial and fine-tuned weights of the selected parameters, respectively, after \( t \) epochs.

Figure~\ref{fig:relative_weight_change}(a) shows that selecting parameters based on the gradient-to-weight ratio, $\rho$, leads to the largest relative weight change, indicating that this metric enables us to effectively prioritize the parameters that undergo meaningful updates relative to their scale.
This observation is consistent with the intended design of our method.
In addition, Figure~\ref{fig:relative_weight_change}(b) shows that the proposed parameter selection method achieves the best loss reduction among all the methods. 
This result verifies that the gradient-to-weight ratio not only identifies parameters that are significantly updated, but also selects those that contribute more directly and effectively to optimization progress. These empirical results align well with the theoretical interpretation presented earlier, which showed that the gradient-to-weight ratio captures both the relative scale of parameter updates and their contribution to loss reduction.

\subsection{Entropy-Guided Layer-wise Parameter Selection}
\subsubsection{Strength- and Distribution-Aware Parameter Allocation.}
Here, we propose a principled approach to quantifying layer importance by analyzing the distribution of parameter-wise gradient-to-weight ratios in each layer.
Our approach allocates the number of tunable parameters in each layer based on the quantified layer importance.
This method plays a key role in reducing the number of tunable parameters while maintaining high accuracy. 

For a given layer $\ell$, we first normalize the gradient-to-weight ratio scores across its parameters to obtain a probability distribution:
\begin{equation}
p_\ell^{(i)} = \frac{\rho_\ell^{(i)}}{\sum_j \rho_\ell^{(j)}},
\label{eq:normalize-rho}
\end{equation}
where \( i \) indicates a specific parameter, and \( j \) iterates over all the parameters at layer \( \ell \).
Based on this distribution, we compute the entropy of the layer:
\begin{equation}
\mathcal{H}(\boldsymbol{p}_\ell) = -\sum_i p_\ell^{(i)} \log p_\ell^{(i)},
\label{eq:entropy}
\end{equation}
which serves as an indicator of how the learning signal is distributed across the parameters in a layer. A low entropy indicates that the learning signal is concentrated in a small subset of parameters, enabling a more selective tuning strategy. In contrast, a high entropy suggests that the signal is more evenly spread, implying that a larger number of parameters are collectively important for the task.

Here, we define a layer importance metric which reflects both the overall magnitude of updates and the internal distribution of the learning signal as follows:
\begin{equation}
\alpha_\ell = \left\| \boldsymbol{\rho}_\ell \right\|_2 \cdot \mathcal{H}(\boldsymbol{p}_\ell).
\label{eq:layer-importance}
\end{equation}

Here, the norm term captures the total relative update magnitude of the layer, reflecting how strongly the layer contributes to learning. The entropy term scales this magnitude based on how concentrated or spread out the learning signal is across the parameters. If the signal is sharply concentrated (i.e., low entropy), fewer parameters can be selected without compromising learning capacity. If the signal is spread out (i.e., high entropy), more parameters must be tuned to account for the distributed importance.

This entropy-guided importance metric enables layer-specific and efficient allocation of the tuning budget across layers.
After computing the importance scores $\alpha_\ell$ for all tunable layers, we normalize them across layers to form a proportional allocation:
\begin{equation}
\gamma_\ell = \frac{\alpha_\ell}{\sum_j \alpha_j},
\label{eq:importance-norm}
\end{equation}
where $\gamma_\ell$ represents the relative share of the total parameter budget allocated to layer $\ell$. Given the total tunable parameter ratio $r$ and the total number of parameters $N$, we determine the number of selected parameters in each layer as
\begin{equation}
k_\ell = \left\lfloor r \cdot N \cdot \gamma_\ell \right\rfloor.
\label{eq:topk-alloc}
\end{equation}
We then select the top-$k_\ell$ parameters for each layer $\ell$, ranked by their gradient-to-weight ratio scores $\rho_\ell^{(i)}$.

\begin{algorithm}[t]
\caption{GEM (Gradient-to-Weight Ratio and Entropy-guided Masking)}
\label{alg:GEM}
\begin{algorithmic}[1]
\State \textbf{Input:} Pretrained weight $W_0$, gradient $\nabla \mathcal{L}(W_0)$, tuning ratio $r$
\State Identify tunable modules (e.g., query, value projections)
\For{each tunable layer $\ell$}
    \For{each parameter $w_\ell^{(i)}$ in layer $\ell$}
        \State Compute gradient-to-weight ratio $\rho_\ell^{(i)}$ \Comment{Eq.~\eqref{eq:gradient-to-weight}}
    \EndFor
    \State Normalize to get probabilities $p_\ell^{(i)}$ \Comment{Eq.~\eqref{eq:normalize-rho}}
    \State Compute layer entropy $\mathcal{H}(\boldsymbol{p}_\ell)$ \Comment{Eq.~\eqref{eq:entropy}}
    \State Compute layer importance $\alpha_\ell$ \Comment{Eq.~\eqref{eq:layer-importance}}
\EndFor
\State Normalize importance scores across layers: $\gamma_\ell$ \Comment{Eq.~\eqref{eq:importance-norm}}
\For{each layer $\ell$}
    \State Compute number of tunable parameters $k_\ell$ \Comment{Eq.~\eqref{eq:topk-alloc}}
    \State Select top-$k_\ell$ parameters with highest $\rho_\ell^{(i)}$
    \State Construct binary mask $M_\ell$ where $[M_\ell]_{m,n} = 1$ if $w_\ell^{(i)}$ is selected, else $0$
\EndFor
\State \textbf{Output:} Binary masks $M_\ell$ for each tunable layer $\ell$
\end{algorithmic}
\end{algorithm}

\subsubsection{Experimental Analysis of Entropy-aware Layer Score.}
We validate our entropy-guided importance score $\alpha_\ell$ defined in Equation~\ref{eq:layer-importance}. Table~\ref{tab:entropy-results} summarizes the results across three GLUE-style tasks (SST-2, BoolQ, and MultiRC), comparing three layer-wise parameter selection strategies: (1) \textit{Uniform}, where each layer receives the same number of trainable parameters; (2) \textit{Norm only}, which allocates parameters to layers solely based on the magnitude of their gradient-to-weight ratio norm $\|\boldsymbol{\rho}_\ell\|_2$; and (3) \textit{Norm $\times$ Entropy}, our proposed metric $\alpha_\ell$, which jointly considers both the magnitude and the intra-layer distribution of learning signals.

Table \ref{tab:entropy-results} (top) shows the share of the total gradient-to-weight signal each method captures. \textit{Norm only} outperforms the uniform baseline, but adding entropy covers even more signal with the same parameter budget, confirming that intra-layer diversity helps identify more informative parameters.
Table \ref{tab:entropy-results} (bottom) reports validation accuracy for each strategy. Methods that capture more of the gradient-to-weight signal achieve higher accuracy. Our entropy-guided approach beats both \textit{Uniform} and \textit{Norm only} on every task, showing the value of modeling both signal strength and diversity in PEFT.

Based on the two key observations above, we conclude that the proposed entropy-guided layer importance metric, $\alpha_\ell$, is a comprehensive measure for determining how many parameters should be fine-tuned at each layer, consistently selecting more informative parameters.
In particular, our empirical study demonstrates that entropy serves as a useful layer-wise information metric, enabling us to maximize PEFT efficiency and improve downstream performance.

\begin{table}[t]
\centering
\footnotesize
\setlength{\tabcolsep}{10pt} 
\renewcommand{\arraystretch}{1.3}
\begin{tabular}{|l|c|c|c|}
\hline
\textbf{Task} & \textbf{Uniform} & \textbf{Norm only} & \textbf{Norm $\times$ Entropy} \\
\hline
\multicolumn{4}{|c|}{\textit{Captured GWR by Selected Parameters (\% of Total)}} \\
\hline
SST-2    & 49.69\% & 68.25\% & \textbf{71.70\%} \\
BoolQ    & 49.52\% & 71.32\% & \textbf{74.22\%} \\
MultiRC  & 47.67\% & 68.52\% & \textbf{72.36\%} \\
\hline
\multicolumn{4}{|c|}{\textit{Validation Accuracy}} \\
\hline
SST-2    & 91.74\% & 93.11\% & \textbf{94.84\%} \\
BoolQ    & 68.33\% & 69.91\% & \textbf{72.39\%} \\
MultiRC  & 65.08\% & 66.56\% & \textbf{70.31\%} \\
\hline
\end{tabular}
\vspace{6pt}
\caption{
\textbf{Entropy-aware strategy captures more informative parameters across layers.}
\normalfont{Top: share of the total gradient-to-weight magnitude captured by the chosen parameters (how well each method covers important weights under a fixed budget).
Bottom: corresponding validation accuracy.
We trained OPT-1.3B on BoolQ, fine-tuning $0.1\%$ of parameters using the gradient-to-weight ratio as the importance metric.
}}
\label{tab:entropy-results}
\end{table}

\subsection{GEM Framework for Sparse Fine-Tuning}
Algorithm~\ref{alg:GEM} presents pseudo code of our GEM framework built upon the proposed parameter prioritization and selection methods. Here, we define how to build masks for PEFT at each layer.
Let $W_\ell \in \mathbb{R}^{d_{\text{in}} \times d_{\text{out}}}$ denote a weight matrix in layer $\ell$ and $M_\ell \in \{0, 1\}^{d_{\text{in}} \times d_{\text{out}}}$ be the corresponding binary mask. We define $M_\ell$ elementwise as:
\[
[M_\ell]_{m,n} = 
\begin{cases}
1, & \text{if } [W_\ell]_{m,n} \text{ is selected for tuning}, \\
0, & \text{otherwise},
\end{cases}
\]
and apply the mask during weight updates as follows:
\begin{equation}
W_\ell \leftarrow W_\ell - \eta \cdot \nabla_{W_\ell} \mathcal{L} \odot M
\label{eq:masked-update}
\end{equation}
where $\eta$ is the learning rate and $\odot$ denotes element-wise multiplication. This ensures that only the selected parameters receive updates, while the rest remain frozen, enabling sparse and efficient fine-tuning.
\section{Experiments} \label{sec:exp}

\begin{table*}[ht]
\centering
\footnotesize
\setlength{\tabcolsep}{1mm}
\resizebox{1\textwidth}{!}{
\begin{tabular}{llccccccccc}
\toprule
\textbf{Model} & \textbf{Method} & \textbf{Params} & \textbf{RTE} & \textbf{SST-2} & \textbf{WiC} & \textbf{BoolQ} & \textbf{MultiRC} & \textbf{COPA} & \textbf{SQuAD} & \textbf{Avg} \\
\midrule
\multirow{8}{*}{OPT-125m}
& FFT & 100\% & 61.21{\scriptsize$\pm$0.6\%} & 89.33{\scriptsize$\pm$0.5\%} & 60.56{\scriptsize$\pm$1.2\%} & 62.81{\scriptsize$\pm$0.7\%} & 64.35{\scriptsize$\pm$0.7\%} & 69.67{\scriptsize$\pm$0.5\%} & 62.71{\scriptsize$\pm$0.9\%} & 67.23 \\
& LoRA & 0.235\% & 60.17{\scriptsize$\pm$0.8\%} & 89.49{\scriptsize$\pm$0.8\%} & 59.40{\scriptsize$\pm$0.7\%} & 63.17{\scriptsize$\pm$0.6\%} & 64.79{\scriptsize$\pm$1.0\%} & 69.00{\scriptsize$\pm$0.7\%} & 62.70{\scriptsize$\pm$1.1\%} & 66.96 \\
& BitFit & 0.082\% & 61.13{\scriptsize$\pm$0.7\%} & 88.69{\scriptsize$\pm$1.1\%} & 58.31{\scriptsize$\pm$0.7\%} & 63.26{\scriptsize$\pm$0.7\%} & 66.52{\scriptsize$\pm$0.6\%} & 69.33{\scriptsize$\pm$0.4\%} & 57.98{\scriptsize$\pm$0.6\%} & 66.46 \\
& Adapter & 0.25\% & 60.65{\scriptsize$\pm$0.7\%} & 89.07{\scriptsize$\pm$1.2\%} & 60.14{\scriptsize$\pm$1.5\%} & 62.52{\scriptsize$\pm$1.1\%} & 65.53{\scriptsize$\pm$0.9\%} & \textbf{\boldmath69.67}{\scriptsize$\pm$1.2\%} & 60.47{\scriptsize$\pm$0.8\%} & 66.86 \\
& AdaLoRA & 0.246\% & 61.61{\scriptsize$\pm$0.5\%} & 88.76{\scriptsize$\pm$0.8\%} & 58.52{\scriptsize$\pm$0.8\%} & 63.42{\scriptsize$\pm$1.0\%} & 66.59{\scriptsize$\pm$0.6\%} & 69.33{\scriptsize$\pm$0.7\%} & 61.98{\scriptsize$\pm$0.9\%} & 67.17 \\
& Random Mask & 0.1\% & 62.82{\scriptsize$\pm$1.5\%} & 89.33{\scriptsize$\pm$0.9\%} & 59.87{\scriptsize$\pm$0.9\%} & 63.01{\scriptsize$\pm$0.8\%} & 60.95{\scriptsize$\pm$1.3\%} & 66.33{\scriptsize$\pm$1.3\%} & 61.69{\scriptsize$\pm$0.9\%} & 66.29 \\
& Top Gradient Mask & 0.1\% & 51.26{\scriptsize$\pm$0.7\%} & 88.07{\scriptsize$\pm$0.6\%} & 57.68{\scriptsize$\pm$1.0\%} & 62.38{\scriptsize$\pm$0.9\%} & 63.83{\scriptsize$\pm$0.8\%} & 66.33{\scriptsize$\pm$0.9\%} & 61.03{\scriptsize$\pm$0.5\%} & 64.37 \\
& GEM (ours) & 0.1\% & \textbf{65.10}{\scriptsize$\pm$0.9\%} & \textbf{89.53}{\scriptsize$\pm$0.7\%} & \textbf{64.16}{\scriptsize$\pm$1.2\%} & \textbf{63.65}{\scriptsize$\pm$0.9\%} & \textbf{66.89}{\scriptsize$\pm$0.7\%} & \textbf{69.67}{\scriptsize$\pm$0.6\%} & \textbf{62.98}{\scriptsize$\pm$0.7\%} & \textbf{68.85} \\
\midrule

\multirow{8}{*}{OPT-1.3b}
& FFT & 100\% & 73.66{\scriptsize$\pm$0.7\%} & 93.42{\scriptsize$\pm$0.6\%} & 67.13{\scriptsize$\pm$0.9\%} & 72.05{\scriptsize$\pm$0.8\%} & 68.31{\scriptsize$\pm$0.9\%} & 82.33{\scriptsize$\pm$0.6\%} & 81.83{\scriptsize$\pm$0.8\%} & 76.96 \\
& LoRA & 0.120\% & 72.56{\scriptsize$\pm$0.6\%} & 93.65{\scriptsize$\pm$1.2\%} & 63.01{\scriptsize$\pm$0.9\%} & 72.01{\scriptsize$\pm$0.7\%} & 68.90{\scriptsize$\pm$0.4\%} & \textbf{83.00}{\scriptsize$\pm$0.9\%} & 82.22{\scriptsize$\pm$0.8\%} & 76.48 \\
& BitFit & 0.041\% & 72.08{\scriptsize$\pm$0.8\%} & 92.65{\scriptsize$\pm$0.6\%} & 62.07{\scriptsize$\pm$0.6\%} & 71.10{\scriptsize$\pm$0.6\%} & 68.43{\scriptsize$\pm$0.9\%} & 79.67{\scriptsize$\pm$0.4\%} & 80.24{\scriptsize$\pm$0.6\%} & 75.18 \\
& Adapter & 0.127\% & 72.80{\scriptsize$\pm$1.1\%} & 92.34{\scriptsize$\pm$0.8\%} & 63.90{\scriptsize$\pm$0.7\%} & 69.02{\scriptsize$\pm$0.6\%} & 69.57{\scriptsize$\pm$1.2\%} & 79.00{\scriptsize$\pm$0.8\%} & 82.21{\scriptsize$\pm$0.7\%} & 75.55 \\
& AdaLoRA & 0.125\% & 70.16{\scriptsize$\pm$0.7\%} & 93.71{\scriptsize$\pm$0.9\%} & 65.20{\scriptsize$\pm$0.6\%} & 72.52{\scriptsize$\pm$0.9\%} & 68.89{\scriptsize$\pm$0.5\%} & 82.67{\scriptsize$\pm$1.0\%} & \textbf{82.61}{\scriptsize$\pm$0.7\%} & 76.54 \\
& Random Mask & 0.1\% & 69.55{\scriptsize$\pm$1.1\%} & 93.15{\scriptsize$\pm$0.8\%} & 63.22{\scriptsize$\pm$0.8\%} & 69.76{\scriptsize$\pm$1.6\%} & 66.90{\scriptsize$\pm$0.7\%} & 81.33{\scriptsize$\pm$0.7\%} & 81.22{\scriptsize$\pm$1.2\%} & 75.02 \\
& Top Gradient Mask & 0.1\% & 67.75{\scriptsize$\pm$0.9\%} & 91.74{\scriptsize$\pm$0.9\%} & 61.23{\scriptsize$\pm$0.5\%} & 62.17{\scriptsize$\pm$0.8\%} & 59.54{\scriptsize$\pm$1.0\%} & 77.67{\scriptsize$\pm$0.6\%} & 79.83{\scriptsize$\pm$0.8\%} & 71.42 \\
& GEM (ours) & 0.1\% & \textbf{74.73}{\scriptsize$\pm$0.7\%} & \textbf{93.84}{\scriptsize$\pm$0.9\%} & \textbf{65.47}{\scriptsize$\pm$0.6\%} & \textbf{72.73}{\scriptsize$\pm$0.9\%} & \textbf{70.31}{\scriptsize$\pm$0.7\%} & \textbf{83.00}{\scriptsize$\pm$0.4\%} & 82.23{\scriptsize$\pm$0.8\%} & \textbf{77.47} \\
\midrule

\multirow{8}{*}{\shortstack{Microsoft \\ Phi-2 \\ 2.7b}}
& FFT & 100\% & 83.49{\scriptsize$\pm$0.8\%} & 93.58{\scriptsize$\pm$0.7\%} & 69.75{\scriptsize$\pm$0.7\%} & 84.10{\scriptsize$\pm$0.8\%} & 83.51{\scriptsize$\pm$1.0\%} & 92.00{\scriptsize$\pm$0.0\%} & 90.46{\scriptsize$\pm$0.7\%} & 85.27 \\
& LoRA & 0.094\% &
76.35{\scriptsize$\pm$0.9\%} & 93.06{\scriptsize$\pm$0.9\%} & 66.97{\scriptsize$\pm$0.8\%} & 83.58{\scriptsize$\pm$0.8\%} & 79.17{\scriptsize$\pm$0.7\%} & 90.67{\scriptsize$\pm$0.6\%} & 90.48{\scriptsize$\pm$1.2\%} & 82.90 \\
& BitFit & 0.009\% & 74.73{\scriptsize$\pm$0.9\%} & 92.83{\scriptsize$\pm$0.6\%} & 60.86{\scriptsize$\pm$0.9\%} & 83.27{\scriptsize$\pm$0.8\%} & 78.87{\scriptsize$\pm$0.9\%} & 87.67{\scriptsize$\pm$0.5\%} & 88.52{\scriptsize$\pm$0.5\%} & 80.96 \\
& Adapter & 0.200\% & 74.01{\scriptsize$\pm$0.8\%} & 92.43{\scriptsize$\pm$0.9\%} & 61.88{\scriptsize$\pm$0.5\%} & 83.10{\scriptsize$\pm$0.7\%} & 77.87{\scriptsize$\pm$0.9\%} & 84.33{\scriptsize$\pm$1.1\%} & 89.02{\scriptsize$\pm$0.7\%} & 80.38 \\
& AdaLoRA & 0.138\% & 81.49{\scriptsize$\pm$0.9\%} & 94.04{\scriptsize$\pm$0.8\%} & 67.82{\scriptsize$\pm$0.9\%} & 84.77{\scriptsize$\pm$0.6\%} & 82.92{\scriptsize$\pm$0.6\%} & 91.00{\scriptsize$\pm$0.8\%} & 90.44{\scriptsize$\pm$0.7\%} & 84.64 \\
& Random Mask & 0.1\% & 79.60{\scriptsize$\pm$1.2\%} & 94.09{\scriptsize$\pm$0.9\%} & 68.07{\scriptsize$\pm$0.8\%} & 84.48{\scriptsize$\pm$0.9\%} & 81.96{\scriptsize$\pm$0.9\%} & 90.67{\scriptsize$\pm$0.6\%} & 90.57{\scriptsize$\pm$0.8\%} & 84.21 \\
& Top Gradient Mask & 0.1\% & 77.98{\scriptsize$\pm$0.6\%} & 94.32{\scriptsize$\pm$0.9\%} & 66.11{\scriptsize$\pm$0.8\%} & 83.82{\scriptsize$\pm$0.8\%} & 82.92{\scriptsize$\pm$0.9\%} & 89.00{\scriptsize$\pm$0.7\%} & 89.92{\scriptsize$\pm$0.8\%} & 83.44 \\
& GEM (ours) & 0.1\% & \textbf{83.94}{\scriptsize$\pm$0.7\%} & \textbf{94.44}{\scriptsize$\pm$0.6\%} & \textbf{69.56}{\scriptsize$\pm$0.6\%} & \textbf{85.02}{\scriptsize$\pm$0.9\%} & \textbf{83.46}{\scriptsize$\pm$0.9\%} & \textbf{91.33}{\scriptsize$\pm$0.8\%} & \textbf{90.89}{\scriptsize$\pm$0.7\%} & \textbf{85.52} \\
\bottomrule
\end{tabular}
}
\caption{Performance comparison across various PEFT methods and baselines on different NLP tasks. Accuracy is reported along with the corresponding standard deviation over 3 random seeds.}
\label{tab:entire_results}
\end{table*}

\subsection{Experimental Setup}
\subsubsection{Datasets and Models.}
We evaluate GEM on seven SuperGLUE~\citep{wang2019superglue} and GLUE~\citep{wang2018glue} benchmarks: RTE~\citep{dagan2005pascal}, COPA~\citep{roemmele2011choice}, SST-2~\citep{socher2013sst}, WiC~\citep{DBLP:journals/corr/abs-1808-09121}, BoolQ~\citep{clark2019boolq}, MultiRC~\citep{MultiRC2018}, and SQuAD v2.0~\citep{rajpurkar-etal-2018-know}.
We report accuracy for the classification tasks and F1 for SQuAD v2.0.

We use three pre-trained language models with varying model capacities: OPT-125M, OPT-1.3B~\citep{zhang2022opt}, and Microsoft Phi-2~\citep{phi2}.
Microsoft Phi-2~\citep{phi2} is a larger model with 2.7 billion parameters, known for its strong language modeling performance and reasoning tasks.

\subsubsection{Baseline Methods.}
To evaluate the effectiveness of GEM, we compare it to several representative PEFT methods: LoRA~\citep{hu2021loralowrankadaptationlarge}, BitFit~\citep{zaken2022bitfitsimpleparameterefficientfinetuning}, Adapter~\citep{houlsby2019parameterefficienttransferlearningnlp}, AdaLoRA~\citep{zhang2023adaloraadaptivebudgetallocation}, Random Masking~\citep{xu2024randommaskingfindswinning}, and Top Gradient Masking~\citep{zhang2024gradientbasedparameterselectionefficient, li2025enhancinglargelanguagemodel}. Detailed settings for each method are provided in the Appendix.

\subsection{Performance Comparison}
Table \ref{tab:entire_results} compares the fine-tuning performance of PEFT methods on seven language tasks.
As shown in the \textbf{Avg} column, which reports mean accuracy across all seven tasks, GEM outperforms all other SOTA methods with OPT-125M, OPT-1.3B, and Phi-2, consistently surpassing sparse fine-tuning baselines at comparable tuning ratios.
Notably, GEM surpasses Top Gradient Masking by $4.48\%$ on OPT-125m and $6.05\%$ on OPT-1.3b, and outperforms Random Masking by $2.56\%$ and $2.45\%$, respectively. We also see that Random Masking exhibits a higher standard deviation (up to $1.6\%$) compared to other methods, which indicates unstable training behavior. 
Note that Adapter shows significantly lower performance on Phi-2 compared to OPT models, which may be attributed to its sensitivity to residual scaling mismatch in pre-norm architectures like Phi-2, whereas it achieves reasonable performance on post-norm models such as OPT.

GEM also outperforms LoRA, BitFit, and Adapter on every task except only three (OPT-125m and OPT-1.3b on COPA, and Phi-2 on SQuAD). 
Notably, GEM outperforms AdaLoRA across all model sizes, even though it uses fewer tunable parameters and fixed masks, showing that a well-designed static selection can surpass more complex adaptive schemes.

Our extensive empirical study highlights two key strengths of GEM. First, it demonstrates strong performance across diverse task types—including classification (SST-2, WiC), natural language inference (RTE, CB, COPA), and extractive QA (SQuAD), showing both versatility and robustness.
Second, GEM demonstrates effectiveness across both lightweight (OPT-125M) and mid-sized (Phi-2) models, ranking first on all seven tasks and achieving the highest average score, highlighting its applicability under a range of computational constraints.
Therefore, we conclude that the proposed parameter selection strategy is a practical and effective PEFT method that can be readily applied to real-world applications.

\subsubsection{Comparative Study on Masking Strategies.} Here, we specifically compare our masking strategy against Random Masking and Top Gradient Masking under the same tuning ratio (0.1\%) as reported in Table~\ref{tab:entire_results}. Random Masking selects parameters uniformly at random without any parameter prioritization criteria. While previous work~\cite{xu2024randommaskingfindswinning} demonstrated that such random selection can still yield competitive performance, the absence of an importance-based selection mechanism fundamentally limits its effectiveness. Top Gradient Masking prioritizes parameters solely based on the gradient magnitude, disregarding the weight scale and update direction—factors critical for meaningful parameter prioritization. Furthermore, both Random Masking and Top Gradient Masking adopt a uniform selection scheme across layers, assigning equal number of updating parameters per layer. This layer-agnostic strategy further restricts their ability to allocate updates effectively. 
In contrast, GEM selects the number of tuning parameters per layer based on entropy-guided layer scoring, which enables more effective update allocation and leads to superior performance.

\subsection{Ablation Study and Discussion}
\subsubsection{Comparison of Layer Importance Metrics.}
Table~\ref{tab:importance-strategies} compares different strategies for layer-wise importance estimation.
The in-layer parameter selection criterion is fixed to the gradient-to-weight ratio.
The \emph{Uniform} strategy all layers equally, giving each the same number of tunable parameters.
The \emph{Norm Only} approach scores layers by the $\ell_2$ norm of their gradient-to-weight ratios, capturing update strength.
The \emph{Entropy Only} approach ignores magnitude and instead measures how focused those ratios are within each layer.
Finally, our proposed method, GEM, combines both the norm and entropy, targeting layers whose learning signals are both strong and focused.

GEM outperforms every baseline across tasks and models.
These results show that layer importance is uneven: allocating the same parameter count to each layer wastes tuning budget.
Instead, parameters should scale with each layer’s contribution, considering both signal strength (norm) and signal spread (entropy).
Relying on either one alone fails to capture the complex learning dynamics across layers.
GEM’s superior performance validates that jointly modeling both the magnitude and the concentration of learning signals enables more expressive parameter selection.

\begin{table}[t]
\centering
\footnotesize 
\setlength{\tabcolsep}{3mm}
\renewcommand{\arraystretch}{1.25} 
\begin{tabular}{llcccc}
\toprule
\textbf{Model} & \textbf{Task} & \textbf{Uniform} & \textbf{Norm Only} & \textbf{Entropy Only} & \textbf{GEM} \\
\midrule
\multirow{2}{*}{\shortstack{OPT- \\ 125m}} 
    & SST-2  & 85.34 & 87.12 & 88.17 & \textbf{89.53} \\
    & SQuAD  & 59.52 & 60.76 & 60.69 & \textbf{62.98} \\
\midrule
\multirow{2}{*}{\shortstack{OPT- \\ 1.3b}} 
    & SST-2  & 91.98 & 92.84 & 91.92 & \textbf{93.84} \\
    & SQuAD  & 80.48 & 81.72 & 80.69 & \textbf{82.23} \\
\midrule
\multirow{2}{*}{Phi-2}
    & SST-2  & 93.37 & 92.66 & 92.41 & \textbf{94.44} \\
    & SQuAD  & 88.70 & 89.44 & 87.84 & \textbf{90.89} \\
\bottomrule
\end{tabular}
\vspace{6pt}
\caption{Comparison of different layer importance strategies for parameter allocation.}
\label{tab:importance-strategies}
\end{table}

\begin{figure}[t]
    \centering
    \includegraphics[width=0.5\textwidth]{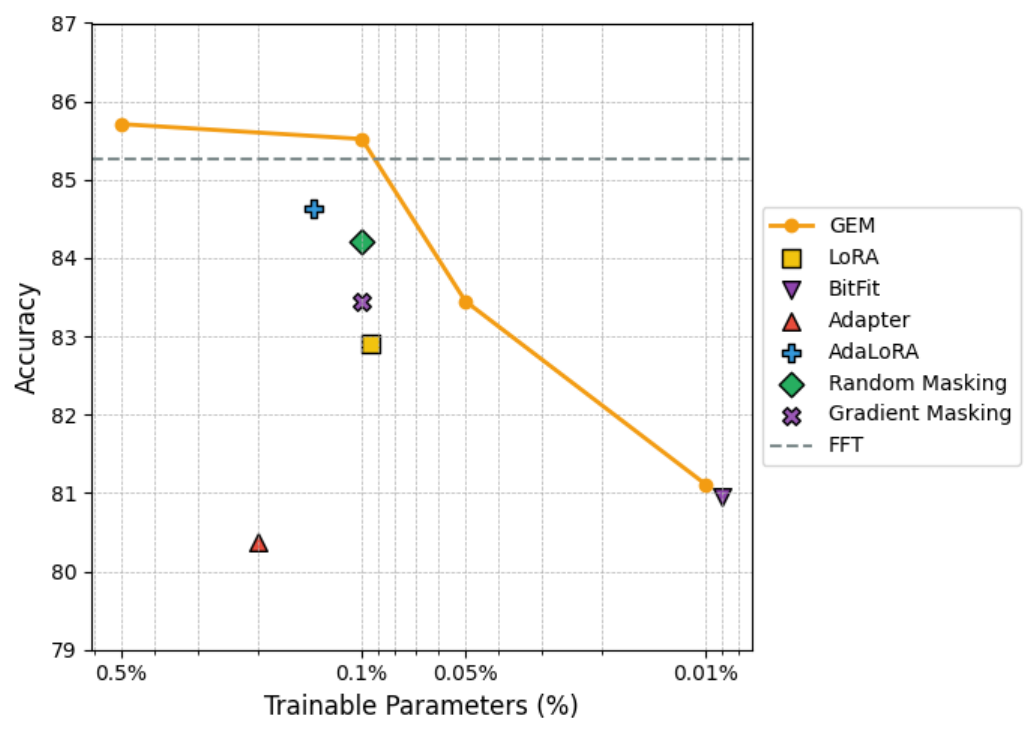}
    \caption{
        Average performance of PEFT methods using Phi-2 on seven tasks.
        GEM shows superior performance over methods with a comparable 0.1\% parameter budget.
    }
    \label{fig:avg_performance}
\end{figure}

\begin{table}[t]
\centering
\footnotesize 
\setlength{\tabcolsep}{3mm} 
\renewcommand{\arraystretch}{1.3}
\begin{tabular}{llccccc}
\toprule
\textbf{Model} & \textbf{Dataset} & \textbf{FFT} & \textbf{LoRA} & \textbf{AdaLoRA} & \textbf{Random} & \textbf{GEM} \\
\midrule
\multirow{2}{*}{Phi-2}
& GSM8k & 53.41 & 42.82 & 41.47 & 47.96 & \textbf{48.78} \\
& MBPP  & 47.83 & 40.24 & 42.94 & 40.18 & \textbf{43.58} \\
\bottomrule
\end{tabular}
\vspace{6pt}
\caption{Evaluation on domain-specific tasks.}
\label{tab:specific-tasks}
\end{table}

\subsubsection{Effect of Varying Tunable Parameter Ratios.} We investigate the performance of GEM under varying tunable parameter ratios: \{0.5\%, 0.1\%, 0.05\%, 0.01\%\}. Figure~\ref{fig:avg_performance} presents the results on Microsoft Phi-2 across seven tasks, alongside baseline methods reported in Table~\ref{tab:entire_results}. 
GEM outperforms all PEFT baselines at roughly the same budget ($0.1\%$).
Performance drops sharply below $0.1\%$, indicating that overly sparse updates cannot adequately adapt the model.
GEM even surpasses full fine-tuning performance when tuning only $0.5\%$ or $0.1\%$ of the model parameters. This observation aligns with recent studies~\cite{xu2024randommaskingfindswinning, li2025enhancinglargelanguagemodel} suggesting that structured and prioritized tuning parameter selection can lead to better generalization than indiscriminate full fine-tuning, which may suffer from overfitting or suboptimal adaptation.

\subsubsection{Transferability to Domain-Specific Tasks.}
Table~\ref{tab:specific-tasks} presents performance comparisons on domain-specific tasks, including mathematical reasoning and code generation. These tasks often require knowledge or reasoning beyond what is captured during pre-training, making them particularly challenging for LLMs to solve without task-specific adaptation. Following \citet{phi2}, we evaluated GSM8k \cite{cobbe2021trainingverifierssolvemath} in a zero-shot setting using the exact match (EM) metric, and MBPP \cite{austin2021programsynthesislargelanguage} with 5-shot prompting using the pass@3 metric.
We exclude OPT-125m and OPT-1.3b from specific task evaluations as these models lack sufficient reasoning and code generation capabilities, resulting in negligible performance.

Across both tasks, all PEFT methods see substantial performance drops relative to full fine-tuning since only about $0.1\%$ of the 2.7-billion parameters in Phi-2 are updated while the model faces strong distribution shifts.
Nevertheless, GEM consistently outperforms other PEFT baselines, demonstrating robustness to distribution shifts. Unlike conventional PEFT methods, GEM prioritizes parameters by their gradient-to-weight ratio, measuring how much each should change relative to its original scale, and updates those with the greatest functional impact. This scale-aware, context-adaptive strategy yields superior performance.

\section{Conclusion} \label{sec:conclusion}

Our paper studies the impact of two key concepts on fine-tuning performance: scale-aware parameter prioritization and distribution-sensitive, layer-wise parameter budget allocation.
We find that the relative weight change is correlated with loss reduction, suggesting that the gradient-to-weight ratio is an effective metric for parameter prioritization.
In addition, our study finds that entropy-guided parameter budget allocation improves fine-tuning performance by effectively allocating tunable parameters based on the parameter value distribution of each layer.
Based on our findings, we develop a practical and effective PEFT framework, GEM, and demonstrate its effectiveness through extensive comparative studies.
We believe the results presented in this study offer a practical fine-tuning strategy for many real-world LLM applications.
An important direction for future work is to develop a theoretical understanding of how gradient-to-weight ratio affects generalization performance.

\appendix

\bibliographystyle{plainnat}
\bibliography{references}
\clearpage
\clearpage
\appendix

\section {Experimental Settings}
\subsection{Dataset Details}

\begin{table}[H]
\centering
\footnotesize
\setlength{\tabcolsep}{3.5mm}
\renewcommand{\arraystretch}{1.25}
\begin{tabular}{lccc}
\toprule
\textbf{\shortstack{Dataset \\ \ }} & 
\textbf{\shortstack{\#Train \\ Samples}} & 
\textbf{\shortstack{\#Evaluation \\ Samples}} & 
\textbf{\shortstack{\#Epochs \\ \ }} \\
\midrule
RTE      & 1,000  & 1,000 & 10 \\
SST-2    & 6,000  & 1,000 & 7  \\
WiC      & 1,000  & 1,000 & 10 \\
BoolQ    & 1,000  & 1,000 & 10 \\
MultiRC  & 1,000  & 1,000 & 10 \\
COPA     & 360    &   40  & 10 \\
SQuAD    & 8,000  & 1,000 & 7  \\
GSM8k    & 2,000  & 1,000 & 10 \\
MBPP     & 600    &  200  & 10 \\
\bottomrule
\end{tabular}
\vspace{6pt}
\caption{Number of training samples, evaluation samples, and training epochs used for each dataset. 
We sample up to a capped number of training samples depending on dataset size, train for a fixed number of epochs, and evaluate on a held-out set of fixed size.}
\label{tab:dataset_usage}
\end{table}

We evaluate our methods on nine datasets, including seven general-domain tasks and two domain-specific tasks, covering classification, question answering, and program synthesis.  
Specifically, we use six general-domain natural language understanding tasks drawn from the GLUE and SuperGLUE benchmarks:  
RTE (textual entailment), SST-2 (sentiment analysis), and WiC (word sense disambiguation) from GLUE,  
as well as BoolQ (yes/no question answering), MultiRC (multi-sentence reading comprehension with multi-label classification),  
and COPA (causal reasoning) from SuperGLUE.
SQuAD is a general-domain question answering dataset.  
GSM8k is a domain-specific mathematical question answering dataset, and MBPP is a domain-specific program synthesis dataset.

The number of training and evaluation samples used for each dataset is summarized in Table~\ref{tab:dataset_usage}.  
For each dataset, we sample a subset of the training set, following the guideline of using either 10\% of the full dataset or up to a maximum cap (1,000 samples for most tasks, 8,000 for SQuAD, and 6,000 for SST-2, 2,000 for GSM8k), depending on the dataset size.  
We report evaluation results on a held-out set of 1,000 samples for evaluation (40 for COPA, 200 for MBPP). 
All experiments are conducted with a batch size of 8.  
We train for 10 epochs on most datasets, except for SST-2 and SQuAD, where we train for 7 epochs due to their larger training sizes.

\subsection{Training Configuration}
Following the practice of Xu et al.~\cite{xu2024randommaskingfindswinning}, we adopt the AdamW optimizer with hyperparameters $\beta_1 = 0.9$, $\beta_2 = 0.999$, and $\epsilon = 1\text{e}{-8}$.  
We use a constant learning rate schedule throughout training and perform a grid search over learning rates in $\{1\text{e}{-2}, 1\text{e}{-3}, 1\text{e}{-4}, 1\text{e}{-5}, 1\text{e}{-6}\}$, each scaled by factors of 1 and 5. The selected learning rates for each method are summarized in Table~\ref{tab:learning_rates_taskwise}.  
All experiments are repeated with three random seeds, and we report the averaged results.

\begin{table*}[t]
\centering
\footnotesize
\setlength{\tabcolsep}{1mm}
\begin{tabular}{llccccccc}
\toprule
\textbf{Model} & \textbf{Method} & \textbf{RTE} & \textbf{SST-2} & \textbf{WiC} & \textbf{BoolQ} & \textbf{MultiRC} & \textbf{COPA} & \textbf{SQuAD} \\
\midrule
\multirow{8}{*}{OPT-125m}
& FFT & 1e-5 & 1e-6 & 1e-6 & 1e-6 & 1e-6 & 5e-6 & 1e-6 \\
& LoRA & 1e-5 & 1e-5 & 5e-5 & 1e-5 & 5e-4 & 1e-5 & 1e-5 \\
& BitFit & 1e-5 & 5e-6 & 1e-5 & 1e-5 & 5e-5 & 1e-5 & 1e-5 \\
& Adapter & 1e-5 & 1e-5 & 1e-5 & 1e-5 & 1e-6 & 1e-5 & 1e-5 \\
& AdaLoRA & 5e-5 & 1e-5 & 1e-5 & 1e-5 & 1e-5 & 5e-5 & 1e-5 \\
& Random Masking & 1e-3 & 1e-2 & 1e-3 & 5e-2 & 1e-3 & 1e-3 & 1e-3 \\
& Top Gradient Masking & 1e-5 & 1e-5 & 1e-5 & 1e-5 & 1e-5 & 1e-5 & 1e-5 \\
& GEM (ours) & 5e-5 & 1e-5 & 1e-5 & 5e-4 & 1e-5 & 1e-5 & 1e-5 \\
\midrule

\multirow{8}{*}{OPT-1.3b}
& FFT & 1e-6 & 1e-6 & 1e-6 & 5e-5 & 1e-6 & 1e-6 & 1e-6 \\
& LoRA & 1e-5 & 1e-5 & 1e-5 & 5e-5 & 1e-5 & 1e-5 & 5e-4 \\
& BitFit & 5e-5 & 1e-5 & 1e-4 & 1e-5 & 5e-4 & 1e-5 & 1e-5 \\
& Adapter & 1e-5 & 1e-4 & 1e-5 & 1e-4 & 1e-5 & 1e-5 & 1e-5 \\
& AdaLoRA & 1e-4 & 1e-5 & 5e-4 & 1e-5 & 1e-5 & 5e-5 & 1e-5 \\
& Random Masking & 1e-2 & 1e-2 & 1e-3 & 1e-3 & 1e-2 & 5e-2 & 1e-3 \\
& Top Gradient Masking & 1e-5 & 5e-5 & 1e-5 & 1e-5 & 5e-4 & 1e-5 & 5e-4 \\
& GEM (ours) & 1e-5 & 5e-5 & 1e-5 & 1e-5 & 5e-4 & 1e-5 & 1e-5 \\
\midrule

\multirow{8}{*}{\shortstack{Microsoft \\ Phi-2 \\ 2.7b}}
& FFT & 5e-5 & 1e-6 & 1e-6 & 1e-6 & 5e-5 & 1e-6 & 1e-6 \\
& LoRA & 1e-5 & 1e-5 & 1e-5 & 1e-5 & 5e-5 & 1e-5 & 5e-5 \\
& BitFit & 1e-5 & 1e-5 & 1e-5 & 1e-5 & 1e-5 & 1e-5 & 1e-5 \\
& Adapter & 1e-5 & 1e-5 & 1e-4 & 1e-5 & 1e-5 & 1e-5 & 1e-5 \\
& AdaLoRA & 1e-5 & 1e-5 & 5e-5 & 1e-5 & 1e-5 & 1e-5 & 5e-5 \\
& Random Masking & 1e-2 & 5e-2 & 1e-3 & 1e-3 & 1e-3 & 1e-3 & 1e-3 \\
& Top Gradient Masking & 1e-5 & 1e-5 & 5e-4 & 1e-5 & 1e-5 & 1e-5 & 1e-5 \\
& GEM (ours) & 1e-5 & 5e-5 & 5e-5 & 1e-5 & 1e-5 & 5e-5 & 1e-5 \\
\bottomrule
\end{tabular}
\caption{Task-specific learning rates selected for each PEFT method across three model scales.}
\label{tab:learning_rates_taskwise}
\end{table*}

\begin{table*}[t]
\centering
\footnotesize
\setlength{\tabcolsep}{1mm}
\begin{tabular}{llcccccccc}
\toprule
Model & Method (Budget) & RTE & SST-2 & WiC & BoolQ & MultiRC & COPA & SQuAD & Avg \\
\midrule
\multirow{4}{*}{\shortstack{Microsoft \\ Phi-2 \\ 2.7B}} 
& GEM (0.5\%)  & 84.02{\scriptsize$\pm$0.8\%} & 94.57{\scriptsize$\pm$0.7\%} & 70.02{\scriptsize$\pm$0.8\%} & 85.18{\scriptsize$\pm$0.8\%} & 83.70{\scriptsize$\pm$0.7\%} & 91.46{\scriptsize$\pm$0.9\%} & 91.03{\scriptsize$\pm$0.8\%} & 85.71 \\
& GEM (0.1\%)  & 83.94{\scriptsize$\pm$0.7\%} & 94.44{\scriptsize$\pm$0.6\%} & 69.56{\scriptsize$\pm$0.6\%} & 85.02{\scriptsize$\pm$0.9\%} & 83.46{\scriptsize$\pm$0.9\%} & 91.33{\scriptsize$\pm$0.8\%} & 90.89{\scriptsize$\pm$0.7\%} & 85.52 \\
& GEM (0.05\%) & 81.67{\scriptsize$\pm$0.5\%} & 92.73{\scriptsize$\pm$1.0\%} & 67.00{\scriptsize$\pm$0.7\%} & 81.34{\scriptsize$\pm$0.6\%} & 81.64{\scriptsize$\pm$0.8\%} & 89.55{\scriptsize$\pm$0.7\%} & 90.24{\scriptsize$\pm$0.7\%} & 83.45 \\
& GEM (0.01\%) & 79.53{\scriptsize$\pm$0.6\%} & 90.56{\scriptsize$\pm$0.4\%} & 65.67{\scriptsize$\pm$0.7\%} & 79.96{\scriptsize$\pm$0.6\%} & 77.02{\scriptsize$\pm$0.9\%} & 87.72{\scriptsize$\pm$0.6\%} & 87.32{\scriptsize$\pm$0.7\%} & 81.11 \\
\bottomrule
\end{tabular}
\caption{Performance of GEM on Microsoft Phi-2 across different tuning budgets. All results are averaged over 3 runs and reported with standard deviation.}
\label{tab:phi2_gem_budget}
\end{table*}

\subsection{Implementation Details for Baseline Methods}

For LoRA, we set the rank to 8 and the scaling factor $\alpha$ to 16.  
For AdaLoRA, we follow its original setup with an initial rank of 12, a target rank of 6, and $\alpha$ set to 16.  
For Adapter, we adopt the original design from \cite{houlsby2019parameterefficienttransferlearningnlp} and configure the bottleneck dimension to 8. 
For Random Masking, we follow \cite{xu2024randommaskingfindswinning} and randomly select the same number of parameters from each layer without any importance criterion. 
Top Gradient Masking selects parameters based on the magnitude of gradients computed from the initial backward pass.  
All fine-tuning methods except BitFit are applied exclusively to the query and value projection matrices.  
BitFit is applied to all bias terms in the network.  
We exclude quantization-based methods from our comparison, as they primarily target compression of large language models (7B--65B) while preserving performance, which differs from our focus on parameter selection strategies under a fixed tuning budget.

\subsection{Computational Budget and Limitation}
All experiments were conducted on two 40GB NVIDIA A100 GPUs. Due to this computational constraint, we were unable to perform experiments on commonly used large-scale models such as 7B models and beyond. Instead, we conducted extensive experiments across a diverse set of datasets to empirically demonstrate the effectiveness of GEM.

\section{Additional Experimental Results and Analysis}

\subsection{Robustness to Dataset Size}

\begin{table*}[t]
\centering
\footnotesize
\setlength{\tabcolsep}{1mm}
\begin{tabular}{llccccccc}
\toprule
\textbf{Model} & \textbf{Task} & \textbf{FFT} & \textbf{LoRA} & \textbf{BitFit} & \textbf{Adapter} & \textbf{Random} & \textbf{Gradient} & \textbf{GEM} \\
\midrule
\multirow{2}{*}{OPT-125m} 
& SST-2  & 91.52  & 91.16 & 89.71 & 90.68 & 90.94 & 90.43 & \textbf{91.47} \\
& SQuAD  & 66.76 & 66.92 & 64.11 & 65.02 & 65.87 & 65.32 & \textbf{67.36} \\
\midrule
\multirow{2}{*}{OPT-1.3b} 
& SST-2  & 95.93 & 95.79 & 95.42 & 95.68 & 94.85 & 91.24 & \textbf{96.09} \\
& SQuAD  & 85.42 & 85.21 & 82.98 & 83.70 & 84.27 & 84.29 & \textbf{85.31} \\
\midrule
\multirow{2}{*}{Microsoft Phi-2} 
& SST-2  & 95.15 & 94.46 & 93.37 & 94.41 & 94.99 & 94.06 & \textbf{95.22} \\
& SQuAD  & 92.02 & 91.74 & 88.64 & 90.16 & 91.81 & 90.76 & \textbf{91.97} \\
\bottomrule
\end{tabular}
\caption{Performance comparison across different PEFT methods using the full training set.}
\label{tab:full_training_set}
\end{table*}

Table~\ref{tab:full_training_set} reports the performance of various PEFT methods when fine-tuned on the full training set of SST-2 and SQuAD. Compared to the results in Table~\ref{tab:entire_results}, which were obtained using limited subsets of the datasets, all methods show expected improvements in accuracy due to access to more training data. In general, fine-tuning on larger datasets requires more tuning parameters to achieve optimal performance, as reflected in the relatively higher accuracy of full fine-tuning compared to other baselines in Table~\ref{tab:entire_results}. We note that GEM continues to outperform all other PEFT baselines and even exceeds the performance of full fine-tuning in some cases when trained on the full dataset, highlighting its robustness to dataset size.

\subsection{Varying Tunable Parameter Ratios}

We investigate how the ratio of tunable parameters affects performance by varying the GEM budget from 0.5\% down to 0.01\%. As shown in Table~\ref{tab:phi2_gem_budget}, increasing the number of tunable parameters generally leads to improved performance across all tasks. Notably, GEM with just 0.1\% of parameters achieves competitive results, with only marginal improvements observed when expanding the budget to 0.5\%. However, when the tuning ratio is reduced to 0.05\% and 0.01\%, the average performance begins to degrade, particularly on tasks such as MultiRC and BoolQ, which appear more sensitive to budget reduction. Despite this, GEM still maintains strong results even under extreme sparsity, demonstrating its robustness and efficiency under limited tuning capacity.

These results highlight GEM's ability to deliver high performance with very few tunable parameters, offering a flexible trade-off between computation cost and accuracy. This table's data is also visualized in Figure~\ref{fig:avg_performance}, which compares the average performance of GEM under varying tuning budgets against other baseline methods.

\begin{figure}[t]
    \centering
    \begin{subfigure}[t]{0.42\linewidth}
        \centering
        \includegraphics[width=\linewidth]{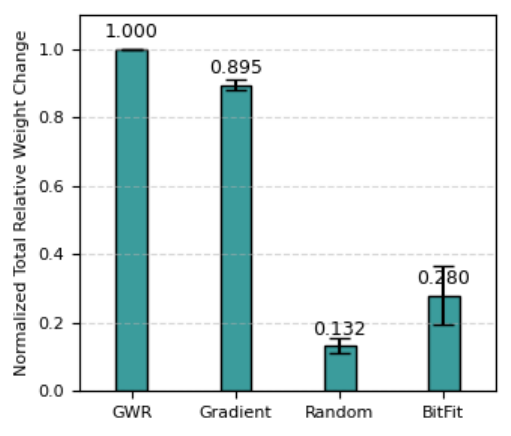}
        \caption{Relative weight change}
        \label{fig:rte_rel_change}
    \end{subfigure}
    \hspace{0.05\linewidth}
    \begin{subfigure}[t]{0.42\linewidth}
        \centering
        \includegraphics[width=\linewidth]{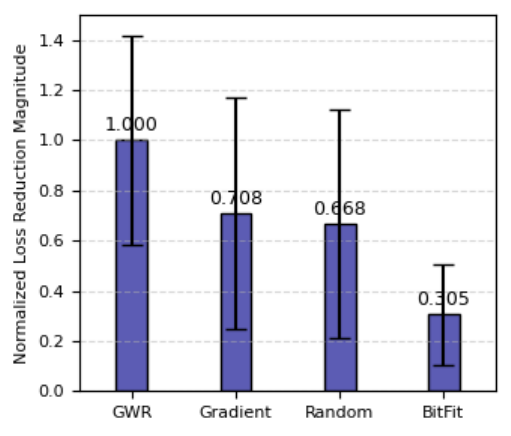}
        \caption{Loss reduction}
        \label{fig:rte_loss_reduction}
    \end{subfigure}
    \caption{
    \textbf{Gradient-to-weight ratio (GWR) consistently highlights critical parameters across different datasets.} \normalfont
    We fine-tune OPT-125M on RTE for 10 epochs, testing four masking strategies over three seeds. Results are normalized.  
    Plot (a) reports total deviation from the pre-trained weights, and plot (b) shows the corresponding loss reduction.  
    Consistent with our main results, gradient-to-weight masking yields both the largest relative weight shifts and the greatest loss drop.
    }
    \label{fig:appendix_relative_weight_change}
\end{figure}

\subsection{Supplementary Experiments on Gradient-to-Weight Ratio}
As discussed in the Figure~\ref{fig:relative_weight_change}, the gradient-to-weight ratio effectively prioritizes parameters that contribute to both significant weight updates and loss reduction.  
Figure~\ref{fig:appendix_relative_weight_change} shows that this trend consistently holds across other datasets as well.

\end{document}